\documentclass[letterpaper, 10 pt, conference]{ieeeconf}

\IEEEoverridecommandlockouts
\usepackage{graphicx}
\usepackage{amsmath, amssymb, amsfonts}
\usepackage{mathtools}
\usepackage{siunitx}
\usepackage{booktabs}
\usepackage{multirow}
\usepackage{tabularx}
\usepackage{hyperref}
\usepackage[capitalize]{cleveref}
\usepackage{xcolor}
\usepackage{algorithm}
\usepackage{algpseudocode}
\usepackage{cite}
\usepackage{graphicx}
\usepackage{subcaption}
\usepackage{tikz}
\usepackage{censor}
\usetikzlibrary{arrows.meta,positioning,fit,calc}

\title{\LARGE\bf
MapForest: A Modular Field Robotics System for Forest Mapping and Invasive Species Localization
}

\author{Sandeep Zachariah$^{1}$, Francisco Yandun$^{1}$, Sachet Korada$^{1}$, and Abhisesh Silwal$^{1}$
\thanks{$^{1}$Robotics Institute, Carnegie Mellon University, Pittsburgh, USA. {
\tt\small sszachar@andrew.cmu.edu}}
}

\begin{document}
\maketitle
\thispagestyle{empty}
\pagestyle{empty}

\begin{abstract}
Monitoring and controlling invasive tree species across large forests, parks, and trail networks is challenging due to limited accessibility, reliance on manual scouting, and degraded under-canopy GNSS. We present MapForest, a modular field robotics system that transforms multi-modal sensor data into GIS-ready invasive-species maps. Our system features: (i) a compact, platform-agnostic sensing payload that can be rapidly mounted on UAV, bicycle, or backpack platforms, and (ii) a software pipeline comprising LiDAR–inertial mapping, image-based invasive-species detection, and georeferenced map generation. To ensure reliable operation in GNSS-intermittent environments, we enhance a LiDAR-inertial mapping backbone with covariance-aware GNSS factors and robust loss kernels. We train an object detector to detect the Tree-of-Heaven \textit{(Ailanthus altissima)} from onboard RGB imagery and fuse detections with the reconstructed map to produce geospatial outputs suitable for downstream decision making. We collected a dataset spanning six sites across urban environments, parks, trails, and forests to evaluate individual system modules, and report end-to-end results on two sites containing Tree-of-Heaven. The enhanced mapping module achieved a trajectory deviation error of 1.95 m over a 1.2 km forest traversal, and the Tree-of-Heaven detector achieved an F1 score of 0.653. The datasets and associated tooling are released to support reproducible research in forest mapping and invasive-species monitoring.

\end{abstract}

\section{Introduction}
Invasive tree species are a growing threat to forest health and biodiversity, and their impacts are amplified when infestations go undetected until they are widespread. Tree-of-Heaven (\textit{Ailanthus altissima}) is one such species: it spreads rapidly, outcompetes native vegetation, and is difficult to eradicate once established. In the northeastern United States, Tree-of-Heaven is also the primary host for the spotted lanternfly, a pest projected to cost Pennsylvania at least \$324 million annually and contribute to the loss of roughly 2,800 jobs if not contained. This economic risk further increases the urgency of early detection and targeted mitigation. Despite this urgency, invasive-species monitoring still relies heavily on manual surveys by field crews. Such workflows are difficult to scale across large parks, trail networks, and forests, and they rarely produce repeatable georeferenced outputs needed for planning and treatment.

Robotics and mobile sensing platforms offer a scalable alternative to manual scouting. They provide high-density, multi-modal data across vast and difficult-to-access terrains that would be impractical to cover extensively for hiking crews. Forest environments pose additional challenges: dense canopy degrades GNSS reliability, uneven terrain that complicates stable and smooth motion, and payload constraints limit the sensing suite that can be deployed on small aerial or human-operated platforms. These conditions demand mapping and perception pipelines that remain robust under intermittent GNSS and that can operate across heterogeneous collection platforms.

To address these challenges, we present \textbf{MapForest}, a modular field robotics system that converts multi-modal sensor streams into georeferenced 3D maps and GIS-compatible layers. The system combines a compact, platform-agnostic sensing payload with a software stack for LiDAR-inertial SLAM, RGB-based invasive-species detection, and geo-tagging of detected trees. We use GLIM~\cite{koide2024glim} as the mapping backbone and extend it with covariance-aware GNSS priors and a robust loss to improve long-range trajectory estimation in dense forests, where under-canopy GNSS is often intermittent and noisy. For invasive-species recognition, we train a YOLOv8~\cite{ultralytics_yolov8} detector to identify Tree-of-Heaven using a curated dataset that combines images collected during our field deployments with additional web-sourced examples. Finally, we fuse detection outputs with the estimated trajectory and map to produce semantic point-cloud visualizations and geospatial layers suitable for downstream GIS analysis and management.

We collected data across six sites spanning urban environments, parks, trails, and forests, and use these datasets to evaluate individual system modules. The contributions of this paper are summarized as follows:
\begin{enumerate}
    \item A compact, platform-agnostic sensing payload and accompanying software stack for scalable forest data collection.
    \item A mapping backbone based on GLIM, extended with robust, covariance-aware GNSS priors to improve long-range trajectory estimation under noisy and intermittent under-canopy GNSS.
    \item An integrated pipeline for Tree-of-Heaven detection and geo-tagging that produces georeferenced 3D maps and GeoTIFF layers, along with a forthcoming public release of data and tooling to support reproducible research.
\end{enumerate}

\section{Related Work}
Automated detection of invasive plant species has attracted growing research interest, with Singh et al. providing a comprehensive review of remote sensing methodologies and best practices for large-scale monitoring~\cite{singh2024systematic}. Within this broader effort, several works target \emph{Ailanthus altissima} specifically. Gao et al.~\cite{gao2025predicting} benchmark deep learning classifiers and vision transformers for Tree-of-Heaven classification from satellite imagery, while Tarantino et al.~\cite{tarantino2019ailanthus} exploit multi-temporal WorldView-2 imagery to map Tree-of-Heaven distribution across seasons. These satellite-scale approaches are valuable for regional assessment and trend analysis, but their reliance on high-resolution commercial imagery can introduce cost and access barriers, and the resulting spatial granularity often limits direct use for site-level treatment planning.

To move from regional mapping to finer-scale sensing, recent work studies UAV imagery for invasive-species detection. Naharki et al.~\cite{naharki2023detection} systematically evaluate RGB, thermal, and NDVI modalities across flight altitudes and seasons for Tree-of-Heaven, providing practical guidance on sensor choice and operational protocols. Related work demonstrates object-detection pipelines for invasive plants in aerial imagery, e.g., a YOLO-based detector for Siam weed~\cite{gautam2025detection}. However, much of this literature reports detection performance primarily in image space. In practice, invasive-species management requires georeferenced outputs such as heatmaps and map layers. Producing these outputs requires additional mapping that are typically outside the scope of detection-centric studies.

Bridging the gap from image-space detections to GIS-ready products motivates a reliable mapping backbone. LiDAR-inertial methods are a standard choice for metric state estimation in large-scale environments where vision can be sensitive to lighting changes, motion blur, and limited texture. Modern tightly-coupled LiDAR-inertial odometry (LIO) systems fuse IMU measurements with scan-to-map registration to handle motion distortion and improve robustness under aggressive motion. FAST-LIO2~\cite{xu2022fast} performs direct registration of raw points to a local map within an iterated Kalman filter framework, enabling real-time operation without explicit feature extraction, while Faster-LIO~\cite{bai2022faster} improves throughput via incremental voxel structures for efficient correspondence search. In contrast, smoothing and mapping approaches such as LIO-SAM~\cite{liosam2020shan} formulate estimation as a factor graph with IMU preintegration and can incorporate loop closure and additional priors (e.g., GNSS) for globally consistent trajectories. GLIM~\cite{koide2024glim} similarly targets globally consistent range--inertial mapping by combining fixed-lag smoothing with keyframe-based scan matching and a global submap optimization backend, leveraging GPU-accelerated scan-matching factors to maintain real-time performance while reducing drift. 

For the perception component, modern object detection methods include efficient single-stage detectors such as YOLO~\cite{redmon2016you}, which enable real-time deployment. Recently, open-vocabulary detectors leveraging vision–language pretraining have been proposed to recognize arbitrary categories specified by text, such as GLIP~\cite{li2022grounded} and Grounding DINO~\cite{liu2024grounding}. While these models offer broad category coverage, in our setting they do not consistently provide the fine-grained accuracy required for reliable species-level detection under real-world appearance variation.

Beyond mapping and detection, integrating above-canopy and under-canopy reconstructions requires robust cross-view alignment. Co-registering aerial (ALS/ULS) and terrestrial (TLS/MLS) forest point clouds is challenging due to strong viewpoint-induced differences and limited point-to-point overlap, which makes correspondence-driven registration unreliable~\cite{besl1992method}. Recent work has emphasized feature-based alignment. Casseau et al.~\cite{casseau2024markerless} proposed a markerless aerial-terrestrial co-registration pipeline that estimates relative constraints between local terrestrial sub-maps and a global, georeferenced aerial map, and then refines alignment via a pose-graph formulation. We use their clique-based horizontal alignment step as a baseline, and refer to it as \emph{Clique} in the remainder of the paper. Castorena et al. (ForestAlign)~\cite{castorena2025forestalign} present a target-less, fully automatic co-registration approach that incrementally aligns scans by progressing from lower-complexity structures (e.g., ground) to higher-complexity structures (e.g., trunks and foliage), iteratively refining alignment for TLS-ALS fusion.
These methods work well in more structured forests (e.g., conifer stands) and in sparse canopies where overlap is limited, but some assumptions can break down in dense, broadleaf forests with heavy canopy occlusion and substantial cross-view appearance changes.


Overall, existing invasive-species monitoring efforts either operate at satellite scale with coarse spatial fidelity or focus on high-resolution sensing without the mapping and fusion needed to produce georeferenced solutions. Our system addresses this gap by coupling ground-level mobile sensing with LiDAR-inertial mapping, RGB-based Tree of Heaven detection, and GIS-ready map generation.

\section{System Overview}
\begin{figure*}[t]
\centering
\includegraphics[width=0.65\textwidth]{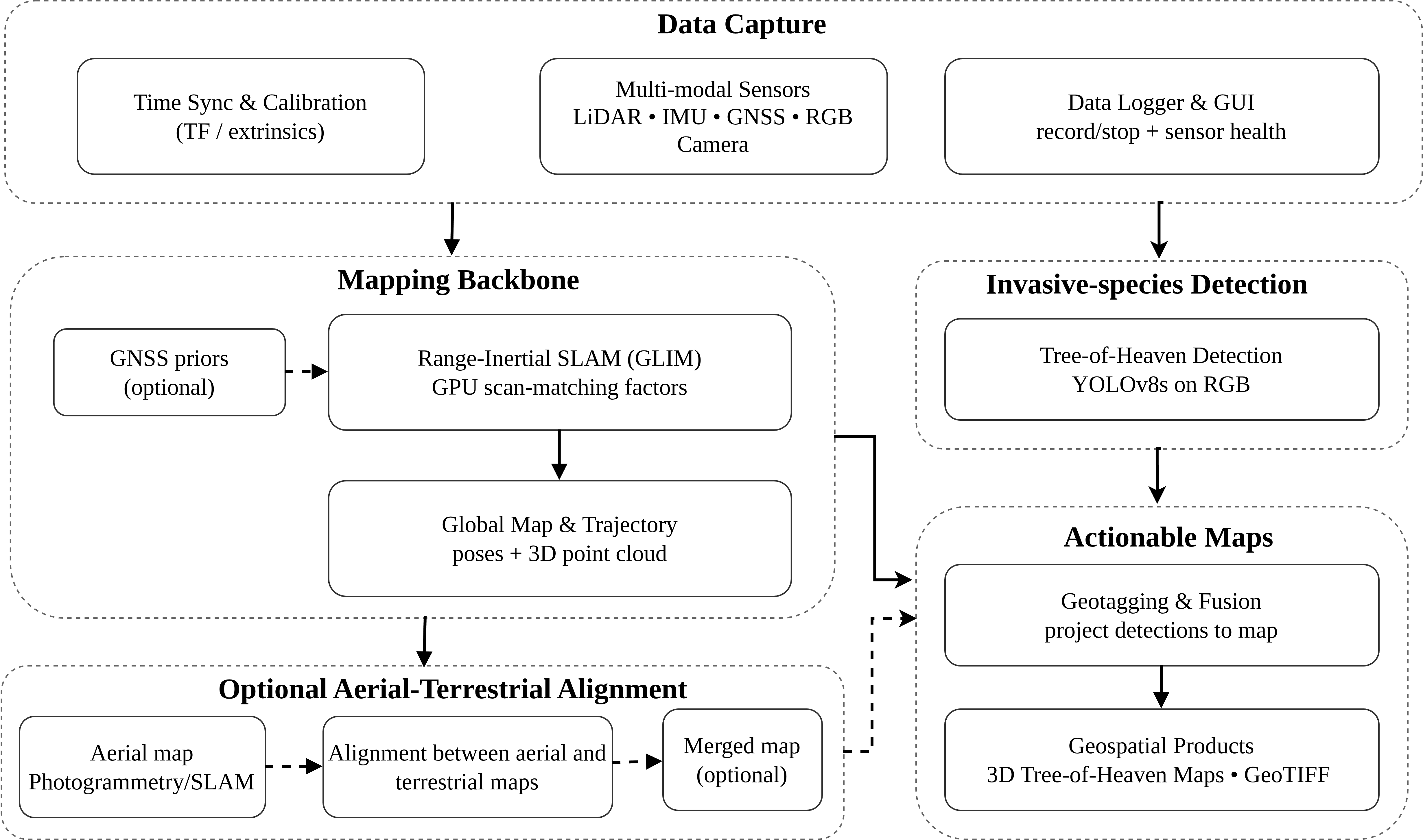}
\caption{System overview. Multi-modal sensing and logging feed a GLIM-based~\cite{koide2024glim} LiDAR-inertial mapping backbone and a YOLOv8 detector. Detections are fused with the estimated trajectory to produce georeferenced map layers (e.g., GeoTIFF/KML). Optional modules include GNSS priors, and aerial-terrestrial map alignment.}
\label{fig:system_overview}
\end{figure*}
Fig.~\ref{fig:system_overview} summarizes our end-to-end pipeline for field-ready forest mapping and invasive-species detection. The system converts time-synchronized multi-modal sensor streams (LiDAR, IMU, GNSS, and RGB) into georeferenced 3D reconstructions and GIS-ready geospatial layers highlighting detected Tree of Heaven.

\textbf{Data capture and synchronization.}
A compact, platform-agnostic payload records LiDAR point clouds, IMU measurements, GNSS observations, and RGB images during field runs. All streams are time-synchronized and transformed into a consistent frame using calibrated extrinsics (Sec.~\ref{sec:hardware_setup}).

\textbf{Mapping backbone.}
Given LiDAR--IMU data, the mapping module estimates the platform trajectory and incrementally builds a globally consistent 3D map. We instantiate this module with GLIM~\cite{koide2024glim} due to its strong performance on long forest traversals and its modular architecture that supports extension to additional constraints and modules and offline refinement of the generated map. To improve long-range consistency under noisy and intermittent under-canopy GNSS, we incorporate covariance-aware GNSS priors with robust loss functions in the pose-graph backend (Sec.~\ref{sec:mapping}).

\textbf{Invasive-species detection and geo-tagging.}
In parallel, the perception module runs a YOLOv8 detector on RGB imagery to identify Tree-of-Heaven instances in each frame. These image space detections are then spatially grounded by projecting them into the reconstructed map using the estimated trajectory and camera-LiDAR calibration. The resulting geotagged detections are used to generate semantic point-cloud overlays and georeferenced raster layers (e.g., GeoTIFF) suitable for downstream GIS analysis (Sec.~\ref{sec:geospatial_grounding}).

\textbf{Optional aerial-terrestrial alignment.}
When aerial reconstructions are available, we optionally align above-canopy and under-canopy maps to obtain a more complete forest representation (Sec.~\ref{sec:map_alignment}).

\section{Hardware Setup}
\label{sec:hardware_setup}
This section describes the sensing payload used for field data collection. The payload is designed for deployment in off-road environments and is guided by practical constraints including platform agnosticism, low power draw, and low cost. These requirements enable affordability and repeatable data collection across diverse carriers and operational contexts.

The payload integrates a Livox Mid-360 LiDAR, GoPro HERO8 RGB camera, Emlid Reach M2 GNSS receiver, and an NVIDIA Jetson Orin Nano for onboard logging. The system is powered by a 3300 mAh battery and operates for approximately 3-4 hours per charge. The payload measures 13 cm $\times$ 13 cm $\times$ 15 cm and weighs 1.1 kg (2.4 lb). Using modular adapters, the payload supports deployment across UAV, bicycle-mounted, handheld, and backpack configurations. To keep the overall system cost low, we use the Mid-360 LiDAR and a commercial off-the-shelf RGB camera rather than higher-cost industrial imaging hardware.

\begin{figure}[h!]
  \centering
  \begin{subfigure}[t]{0.22\linewidth}
    \centering
    \includegraphics[width=\linewidth, trim=0 0 0 5.5cm, clip]{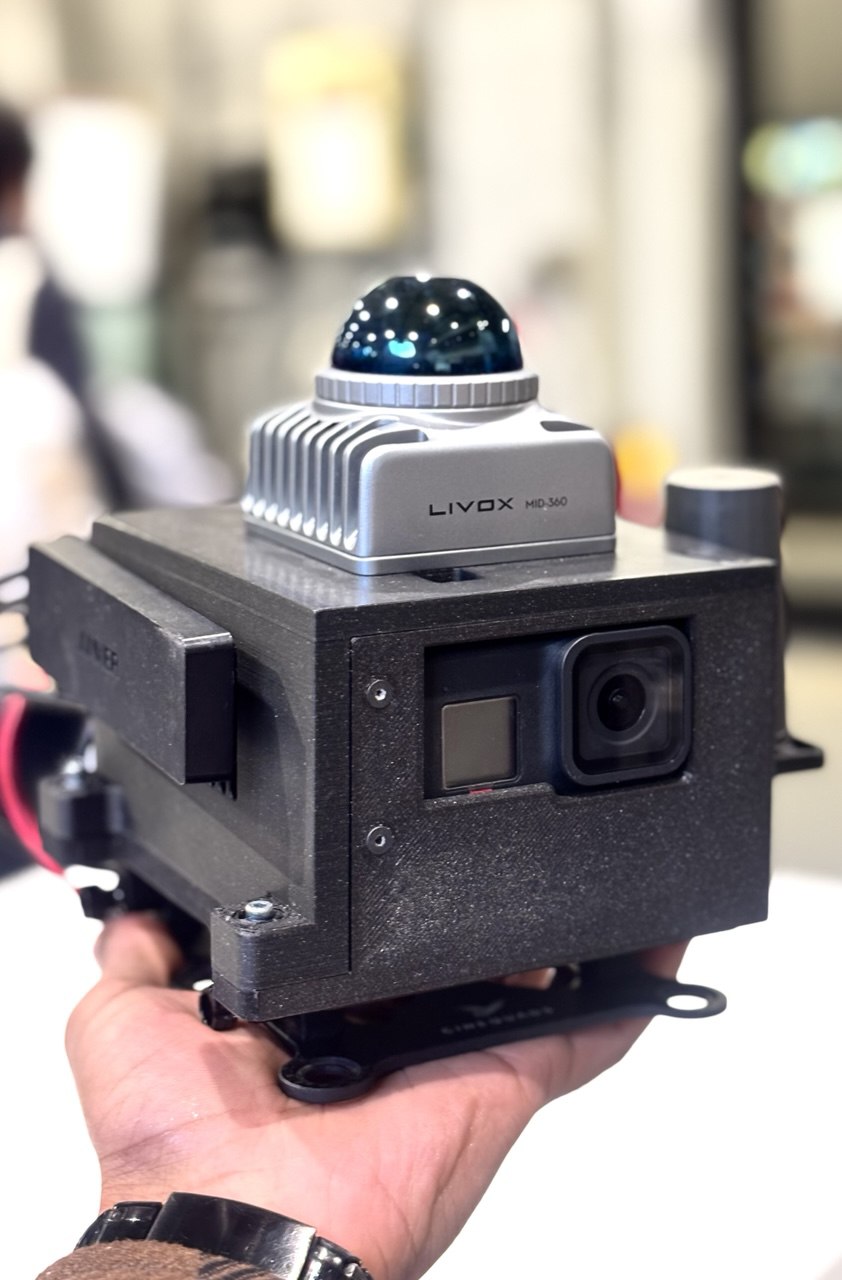}
    \caption{Handheld}
    \label{fig:flagstaff_cov_aware}
  \end{subfigure}
  \begin{subfigure}[t]{0.22\linewidth}
    \centering
    \includegraphics[width=\linewidth]{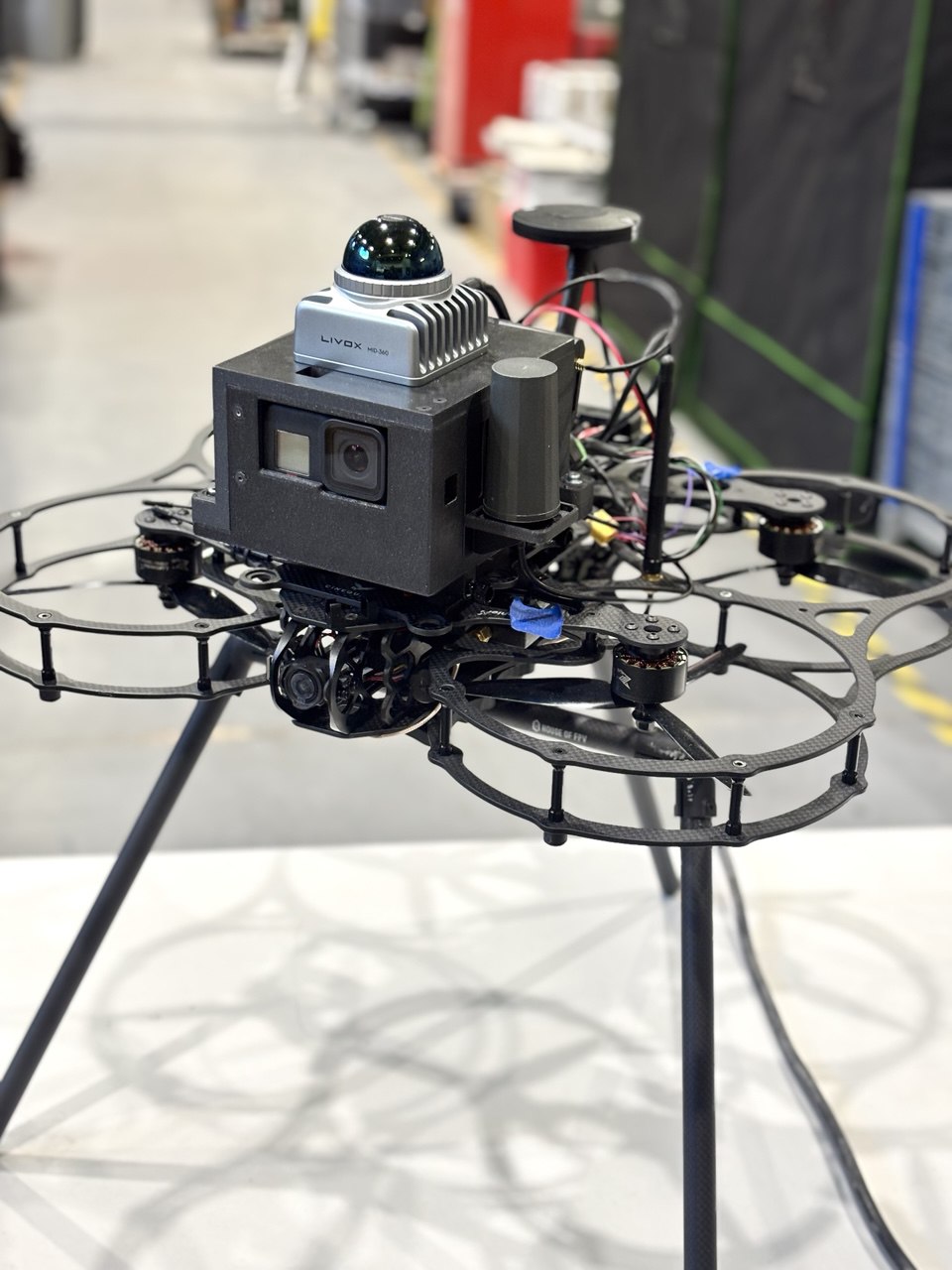}
    \caption{UAV}
    \label{fig:flagstaff_no_gnss}
  \end{subfigure}
  \begin{subfigure}[t]{0.22\linewidth}
    \centering
    \includegraphics[width=\linewidth]{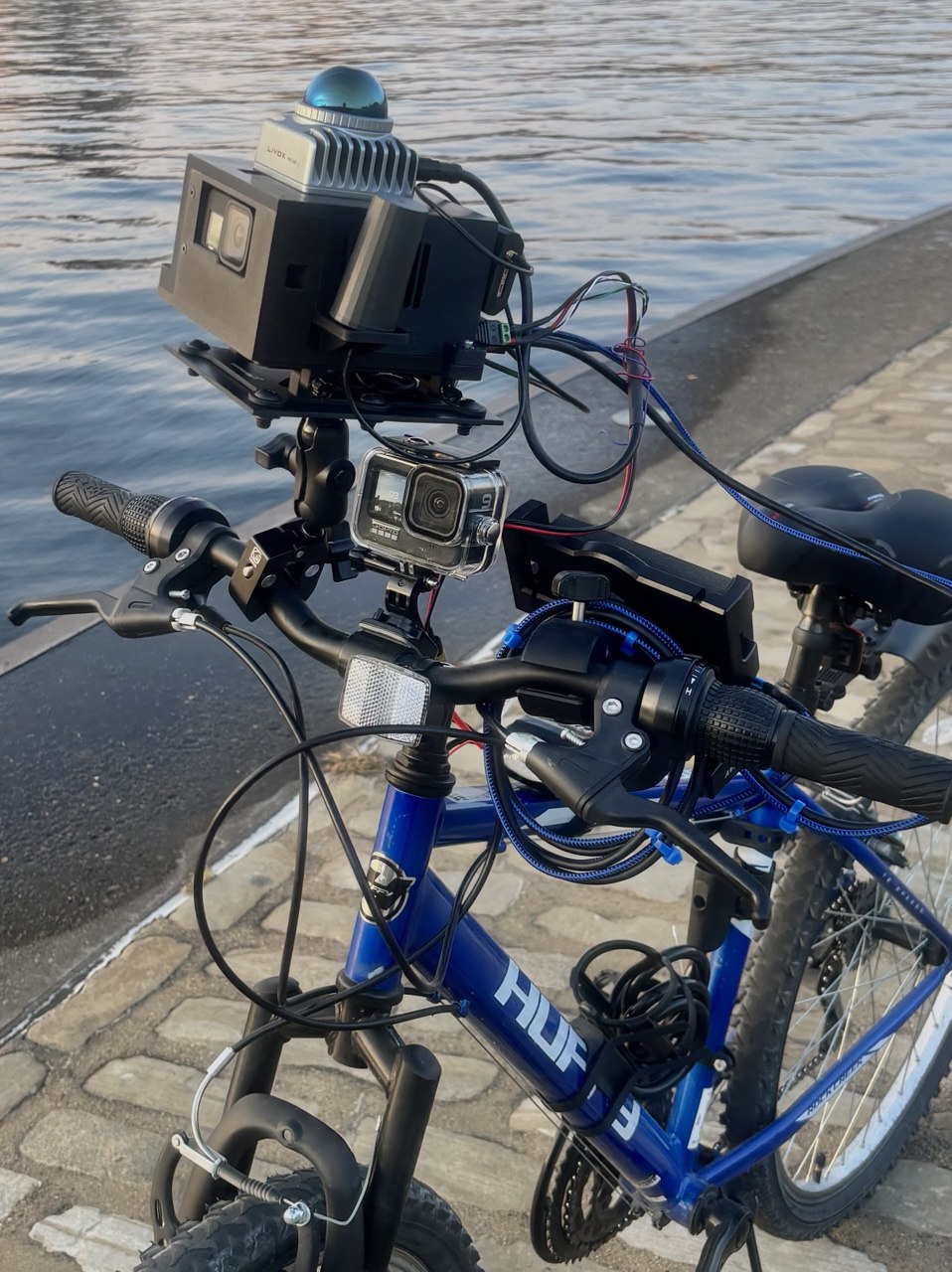}
    \caption{Bicycle}
    \label{fig:flagstaff_cov_aware}
  \end{subfigure}
  \caption{Sensor payload deployed across multiple carriers, demonstrating platform-agnostic operation.}
  \label{fig:sensor_carriers}
\end{figure}

Accurate spatial association between modalities requires repeatable calibration. We use a LiDAR with an integrated IMU (Livox), calibrate the camera intrinsics using the ROS2 camera calibration tools, and estimate camera–LiDAR extrinsics using the lidar\_camera\_calib package~\cite{yuan2021pixel}. Additionally, to improve ease of use in the field, we developed a GUI that enables one-click start/stop data recording and provides real-time sensor health monitoring. The GUI also includes a bag-validation utility that verifies required topics were recorded at expected frequencies, reducing the likelihood of unusable field runs due to silent sensor or driver failures.

\section{Mapping}
\label{sec:mapping}
Accurate mapping and pose estimation are central to producing consistent 3D reconstructions and enabling downstream tasks such as geo-tagging and inventory. Given time-synchronized LiDAR–IMU data, the SLAM module estimates the payload trajectory and incrementally constructs a map of the environment. The resulting map provides a common spatial reference for projecting detections into 3D and for estimating forestry-relevant quantities.

We evaluated multiple LiDAR–inertial SLAM implementations (Fast\_LIO~\cite{xu2022fast}, Faster-LIO~\cite{bai2022faster}, fast\_LIMO, and GLIM~\cite{koide2024glim}). While several approaches perform well in structured outdoor scenes, we found that long forest trails and environments with limited man-made structure can lead to reduced robustness over extended traversals. In our deployments, GLIM provided reliable performance in these settings and offered a modular architecture that supports additional constraints. In particular, its support for manual loop-closure correction and extensibility for integrating external experimental modules, such as GNSS-based factors made it a practical and scalable choice for us.

Because under-canopy GNSS is often noisy and intermittent, we apply a robust loss to GNSS factors to reduce sensitivity to outliers. Robust loss functions (M-estimators) replace the standard squared residual with a robust cost function $\rho(\cdot)$, thereby down-weighting large residuals. In our implementation, we use the Huber loss~\cite{huber1973robust}, which empirically provided the best performance in our setting. The Huber loss is defined as,
\begin{equation}
\rho(r)=
\begin{cases}
\frac{1}{2}r^{2}, & \text{if } |r|\le \delta,\\[4pt]
\delta\left(|r|-\frac{1}{2}\delta\right), & \text{otherwise.}
\end{cases}
\label{eq:huber}
\end{equation}
where $r$ denotes the residual and $\delta>0$ is the threshold that sets the transition between the quadratic and linear region.

Beyond geometric reconstruction, LiDAR intensity provides an additional measurement channel that can be useful for downstream processing. We extend the baseline GLIM mapping pipeline to preserve and propagate per-point intensity through the mapping stage, so the final reconstructed map retains intensity values alongside 3D geometry. This capability is required for our ground-truthing workflow: reflective patches placed on selected trees produce high-intensity returns, enabling reliable association of marked trees in the reconstructed point cloud for geometry-correctness evaluation. This intensity-preservation extension has been integrated into the GLIM core codebase, and we evaluate its impact in Sec.~\ref{sec:experiments}.

\section{Invasive Species Detection}
\label{sec:detection}
We formulate Tree-of-Heaven (\emph{ToH}) detection as a binary object detection problem on RGB imagery, predicting axis-aligned 2D bounding boxes and labels in {\emph{ToH}, \emph{non-ToH/background}}. For training the detector, we use RGB images from two sites (Gascola and Hazelwood) and augment them with additional ToH examples from iNaturalist~\cite{inaturalist}, a public citizen-science biodiversity platform. The iNaturalist images provide higher-resolution close-up views that increase appearance diversity.

We curate a labeled dataset of 1{,}644 images. Table~\ref{tab:toh_dataset_splits} summarizes the dataset composition and the train/validation/test split across sources. The Hazelwood subset was collected using the bicycle-mounted configuration and includes challenging conditions such as motion blur, while Gascola includes partial occlusions and cluttered backgrounds typical of dense trails.

Ground-truth annotations were created in Label Studio by manually drawing bounding boxes and exporting labels in YOLO format. We trained a YOLOv8s~\cite{ultralytics_yolov8} detector for 200 epochs (input resolution 832) with a base learning rate of 0.001 on an RTX 2070 GPU. During inference, the model runs at an average of 13 ms per image on an RTX 2070 GPU.
\begin{table}[h!]
  \centering
  \caption{Tree-of-Heaven detection dataset composition and splits (images).}
  \label{tab:toh_dataset_splits}
  \vspace{-0.6em}
  \begin{tabular}{lccc}
    \toprule
    \textbf{Source} & \textbf{Train} & \textbf{Val} & \textbf{Test} \\
    \midrule
    iNaturalist (web) & 700 & -- & -- \\
    Gascola (field)                & 436 & 54 & 56 \\
    Hazelwood (field)              & 330 & 31 & 32 \\
    \midrule
    \textbf{Total}                 & 1466 & 85 & 88 \\
    \bottomrule
  \end{tabular}
  \vspace{-0.9em}
\end{table}

\section{Geospatial Grounding}
\label{sec:geospatial_grounding}
This section describes the geo-tagging module, which converts image-based detections into physically grounded, georeferenced map layers. Given calibrated camera intrinsics and camera--LiDAR extrinsics, we associate 2D detector outputs with the reconstructed 3D map by projecting image pixels within each detection into the LiDAR frame and assigning the corresponding labels to nearby LiDAR points. This produces a semantic point-cloud overlay (visualized as a 3D heatmap) in the map frame, where points associated with Tree of Heaven are highlighted.

To georeference these outputs, we anchor the reconstructed map to a global coordinate system using the GNSS measurements collected during the data collection run. The resulting geo-referenced map enables export of GIS-compatible layers. In addition to annotated point-cloud maps, we generate GeoTIFF raster layers that can be loaded directly in standard geospatial software (e.g., QGIS), providing map layers that users and practitioners can review and use for treatment planning.

\section{Aerial-Terrestrial Map Alignment}
\label{sec:map_alignment}
Forests are unstructured environments with strong occlusions, spatially varying vegetation density, and limited persistent structure, making it challenging to obtain a complete and consistent representation from a single sensing perspective. Under-canopy mapping captures understory structure and trunk-level geometry, but provides limited visibility of crowns. In contrast, above-canopy mapping captures canopy geometry and crown structure, but contains little information about the understory. Fusing these complementary reconstructions can yield a more complete forest representation, but requires robust alignment between the aerial and terrestrial maps.

We decompose aerial-terrestrial alignment into (i) \emph{horizontal alignment}, which estimates planar translation and yaw via an $\mathrm{SE}(2)$ transform $\mathbf{T}(\boldsymbol{\theta}) \in \mathrm{SE}(2)$ with parameters $\boldsymbol{\theta}=[t_x,\, t_y,\, \psi]^\top$, and (ii) \emph{vertical alignment}, which accounts for the remaining degrees of freedom. Vertical alignment can be obtained by gravity alignment from IMU priors and, when available, aligning a shared ground reference from terrain observations. Horizontal alignment is typically the more challenging component in dense forests because the two reconstructions exhibit minimal point-to-point overlap, which makes correspondence-based registration unreliable.

To address these challenges, we estimate horizontal alignment by posing it as an optimization problem that maximizes the statistical dependence between two likelihood fields derived independently from the aerial and terrestrial reconstructions. Specifically, we construct an aerial field from canopy structure and a terrestrial field from trunk evidence, and then optimize a normalized mutual-information~\cite{maes2003medical} objective to recover the $\mathrm{SE}(2)$ transform. The remainder of this section details the likelihood-field construction and the mutual-information-based optimization procedure.

\subsection{Likelihood-Field Representation}
We adopt a probabilistic formulation that avoids explicit feature correspondences. In dense forests, heavy canopy overlap and viewpoint-induced differences make discrete feature extraction unreliable and increase the risk of spurious correspondences. Instead, we represent tree evidence as continuous likelihood fields that capture the spatial distribution of crowns on the horizontal plane, providing a representation that is robust to sparse or ambiguous features. Specifically, we convert both maps into 2D \emph{tree-likelihood fields} and estimate alignment by maximizing the statistical dependence between these fields.

\paragraph{Aerial likelihood field}
From the aerial reconstruction, we compute a canopy height model (CHM) on a 2D grid, denoted $C(\mathbf{x})$ for $\mathbf{x}=[x,y]^\top$.
We then apply a multi-scale Laplacian-of-Gaussian (LoG) operator to emphasize crown-scale extrema:
\begin{equation}
L_A(\mathbf{x})
= \max_{\sigma \in \mathcal{S}}
\left|
\left(\nabla^2 G_{\sigma}\right) * C
\right|(\mathbf{x}),
\label{eq:aerial_log}
\end{equation}
where $G_{\sigma}$ is a Gaussian kernel with scale $\sigma$, $\nabla^2$ is the Laplacian, $*$ denotes convolution, and $\mathcal{S}$ is a set of crown-relevant scales. Intuitively, regions with pronounced local canopy structure produce larger LoG responses and correspond to high likelihood of tree presence.

\paragraph{Terrestrial likelihood field}
From the terrestrial map, we extract trunk hypotheses (e.g., via trunk segmentation) and represent them as a set of 2D trunk locations $\{\mathbf{p}_i\}_{i=1}^{N}$ in the terrestrial frame.
We form a continuous likelihood field via kernel density estimation (KDE):
\begin{equation}
L_T(\mathbf{x})
= \sum_{i=1}^{N} \exp\!\left(-\frac{\|\mathbf{x}-\mathbf{p}_i\|^2}{2h^2}\right),
\label{eq:terrestrial_kde}
\end{equation}
where $h$ is a bandwidth parameter controlling spatial smoothing. This yields a smooth field that retains trunk spatial structure while remaining robust to missing or noisy detections. In practice, we normalize both fields to $[0,1]$ before computing the alignment objective.

\subsection{Alignment via Normalized Mutual Information}
Given the two likelihood fields, we estimate $\boldsymbol{\theta}$ by maximizing \emph{normalized mutual information} (NMI) between $L_A$ and a transformed version of $L_T$. For each candidate transform $\mathbf{T}(\boldsymbol{\theta})$, we warp the terrestrial field into the aerial frame:
\begin{equation}
\tilde{L}_T(\mathbf{x};\boldsymbol{\theta})
= L_T\!\left(\mathbf{T}(\boldsymbol{\theta})^{-1}\mathbf{x}\right).
\label{eq:warp}
\end{equation}

\begin{figure}[t!]
  \centering

  \includegraphics[width=\linewidth]{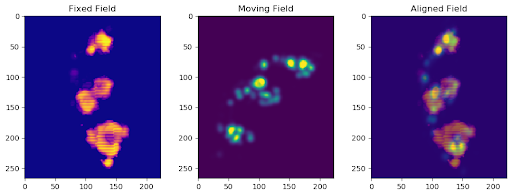}

  \begin{subfigure}[t]{0.32\linewidth}
    \centering
    \includegraphics[width=0.99\linewidth, trim=0 0.5cm 0 2.5cm, clip]{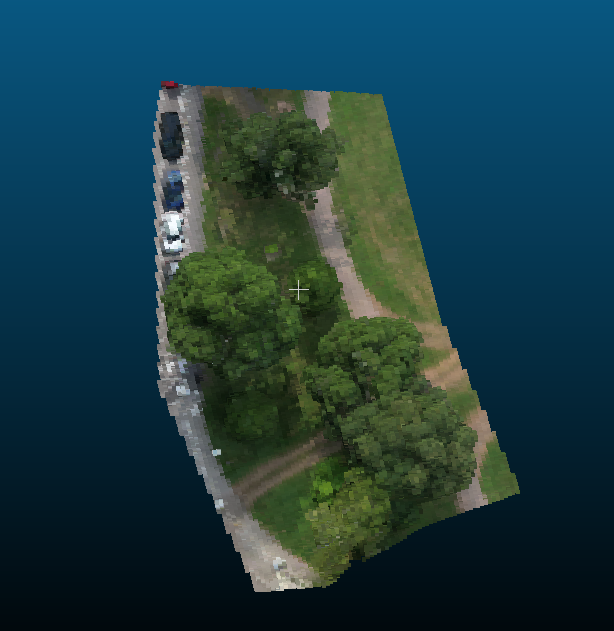}
    \caption{Aerial map}
    \label{fig:acm_field}
  \end{subfigure}
  \begin{subfigure}[t]{0.32\linewidth}
    \centering
    \includegraphics[width=0.99\linewidth, trim=0 0cm 0 0.5cm, clip]{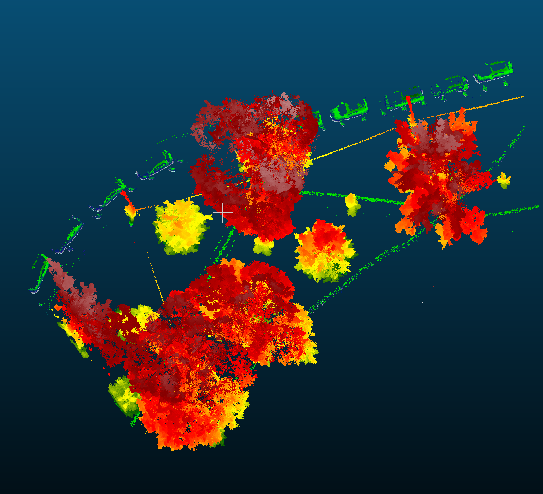}
    \caption{Terrestrial map}
    \label{fig:ucm_field}
  \end{subfigure}
  \begin{subfigure}[t]{0.32\linewidth}
    \centering
    \includegraphics[width=0.99\linewidth, trim=0 0cm 0 0.5cm, clip]{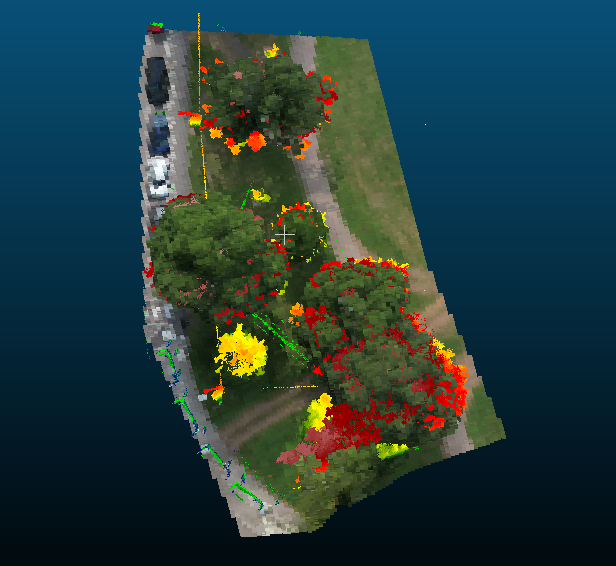}
    \caption{Aligned maps}
    \label{fig:aligned_fields}
  \end{subfigure}

  \caption{Likelihood fields used for aerial–terrestrial alignment. Top row: likelihood fields for (a) aerial, (b) terrestrial, and (c) aligned results. Bottom row: the corresponding point-cloud maps for each case.}
  \label{fig:likelihood_fields}
  \vspace{-0.6em}
\end{figure}

The alignment problem is then posed as:
\begin{equation}
\boldsymbol{\theta}^{*}
=
\arg\max_{\boldsymbol{\theta}\in\Theta}
\ \mathrm{NMI}\!\left(L_A,\tilde{L}_T(\cdot;\boldsymbol{\theta})\right),
\label{eq:opt}
\end{equation}
where $\Theta$ is a bounded search region.
Similar to prior approaches, we assume the two maps are approximately co-registered using GNSS to initialize the search, and restrict $\Theta$ accordingly. Figure~\ref{fig:likelihood_fields} visualizes the fixed (aerial) and moving (terrestrial) likelihood fields, along with the aligned field after optimization, showing strong spatial agreement between the two.

\subsection{Multi-start Optimization}
Because NMI is computed via histogram-based entropy estimates, the objective in~\eqref{eq:opt} is generally non-convex and can exhibit local optima. To improve robustness, we employ a multi-start optimization approach: first sample a set of initializations $\{\boldsymbol{\theta}_k\}_{k=1}^{K}$ around the GNSS prior and run a local optimizer from each, selecting the solution with maximum objective value.

This strategy is a standard practical approach for mitigating poor local convergence in non-convex objectives. To make multi-start tractable, we batch candidate transforms and compute NMI in parallel on the GPU using PyTorch tensor operations and CUDA acceleration; we refer to this implementation as \textit{Tensor-MI}.

\section{Experiments}
\label{sec:experiments}
\subsection{Testing Sites}
We evaluate MapForest on data collected across six sites spanning urban environments, parks, bike-trails, and forests. Not all sites contain Tree-of-Heaven; however, the full set provides diverse operating conditions to evaluate individual system components, including mapping, detection, and geospatial grounding.

We collected data over about one year using the payload in handheld, backpack-mounted, and bicycle-mounted configurations. Across all runs, we record time-synchronized LiDAR, IMU, GNSS, and RGB streams.
\begin{table}[h!]
  \centering
  \caption{Summary of evaluation sites and data collection runs.}
  \label{tab:site_summary}
  \vspace{-0.6em}
  \begin{tabular}{l c c c}
    \toprule
    \textbf{Site} & \textbf{ToH present} & \textbf{Traversal length (km)} & \textbf{RTK GNSS} \\
    \midrule
    Gascola   & Yes & 1.40 & Yes \\
    Bike Trail & Yes & 24.10 & Yes \\
    Chatham   & Yes & 1.20 & Intermittent \\
    Flagstaff & No & 0.80 & Yes \\
    Frick Park & No & 2.22 & Intermittent \\
    CMU Campus & No & 0.48 & No \\
    \bottomrule
  \end{tabular}
  \vspace{-0.9em}
\end{table}

\begin{figure}[h!]
    \centering
    \begin{subfigure}[t]{0.49\linewidth}
    \centering
    \includegraphics[width=0.8\linewidth]{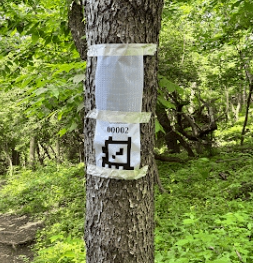}
  \end{subfigure}
  \begin{subfigure}[t]{0.49\linewidth}
    \centering
    \includegraphics[width=0.8\linewidth, trim=0 0 0 0.4cm, clip]{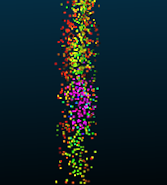}
  \end{subfigure}
    \caption{Marked tree used for DBH evaluation. (Left) AprilTag provides a unique ID and reflective tape yields high-intensity LiDAR returns. (Right) Point cloud of the corresponding tree trunk; the pink region indicates high-intensity returns from the reflective tape.}
    \label{fig:dbh_tree}
\end{figure}

For trajectory evaluation, we use RTK-enabled GNSS recorded during data collection (when available) as the ground-truth trajectory. To assess geometric reconstruction quality, we mark a subset of trees with AprilTags and reflective tape so they can be reliably identified in the reconstructed point cloud. For these marked trees, we record the Diameter at Breast Height (DBH) measurements, which serve as ground truth for evaluating DBH estimates derived from the reconstructed map. The reflective material produces high-intensity returns in LiDAR scans, enabling tree correspondence for DBH evaluation as shown in Fig. \ref{fig:dbh_tree}.

\subsection{Results}
\subsubsection{Mapping Accuracy}
Trajectory estimation performance is quantified using the Absolute Trajectory Error (ATE) between the estimated SLAM trajectory and the RTK-GNSS ground-truth trajectory. We report the error as mean $\pm$ standard deviation to capture both average drift and variability over the traversal. For this comparison, all methods operate in a pure LiDAR--IMU setting (i.e., no GNSS factors are used).

Tab.~\ref{tab:rmse_sites} reports the ATE for the Flagstaff site. Among the evaluated methods, GLIM achieves the lowest error, with $8.67 \pm 4.77$~m over an $0.8$~km traversal. In more challenging sites (e.g., Gascola), we observed that FastLIO, FasterLIO, and FastLIMO frequently diverge after a short duration due to reduced geometric constraints and vegetation-induced degeneracies, preventing successful map completion. In contrast, GLIM consistently produced stable trajectories and complete maps across all six sites. Because competing methods failed to produce valid trajectories on several sites, we report quantitative trajectory error only where all methods successfully completed the run.
\begin{table}[h!]
  \centering
  \caption{Trajectory error (m) w.r.t.\ RTK GNSS ground truth on the Flagstaff site.}
  \label{tab:rmse_sites}
  \vspace{-0.6em}
  \begin{tabular}{l c}
    \toprule
    \textbf{Method} & \textbf{Flagstaff}\\
    \midrule
    FastLIO   & $15.77 \pm 13.91$\\
    FasterLIO & $12.83 \pm 9.02$ \\
    FastLIMO  & $18.70 \pm 11.59$ \\
    GLIM      & $8.67 \pm 4.77$\\
    \bottomrule
  \end{tabular}
  \vspace{-0.9em}
\end{table}

To assess geometric map quality, we evaluate the accuracy of Diameter at Breast Height (DBH) estimates derived from the reconstructed 3D point cloud. Specifically, we compute the mean relative error (MRE) between the estimated DBH values and the tape-measured ground-truth DBH for a set of trees (Eq.~\eqref{eq:mre_dbh}). DBH is estimated using the open-source Forest Structural Complexity Tool (FSCT), which takes the reconstructed map as input and outputs standard forestry attributes (e.g., DBH, height, and volume of trees).

\begin{figure}[h!]
    \centering
    \includegraphics[width=0.7\linewidth]{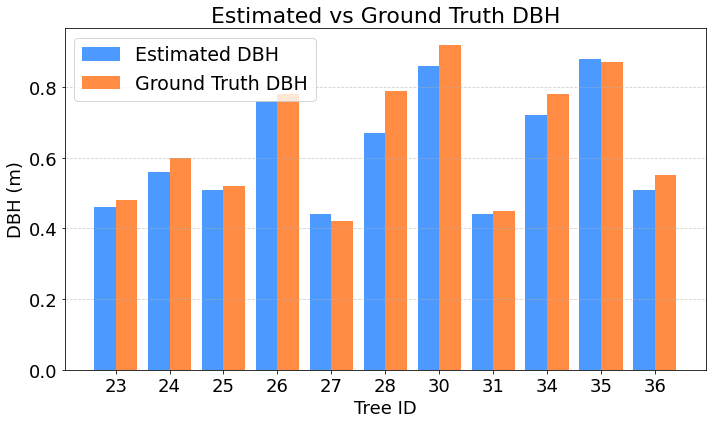}
    \caption{Estimated vs. ground-truth DBH for 11 Flagstaff trees (FSCT on reconstructed map vs. tape-measured).}
    \label{fig:dbh_est_gt}
\end{figure}

On the Flagstaff dataset, we obtain an MRE of \textbf{0.04} across 11 marked trees. Fig.~\ref{fig:dbh_est_gt} compares the FSCT-derived DBH estimates against ground truth for these trees.
\begin{equation}
    \mathrm{MRE}_{\mathrm{DBH}}
    = \frac{1}{N} \sum_{i=1}^{N}
    \frac{\bigl|\mathrm{DBH}^{\mathrm{est}}_i - \mathrm{DBH}^{\mathrm{gt}}_i\bigr|}
    {\mathrm{DBH}^{\mathrm{gt}}_i}
    \label{eq:mre_dbh}
\end{equation}

\subsubsection{GNSS Integration Ablation}
\begin{figure}[h!]
  \centering
  \begin{subfigure}[t]{0.5\linewidth}
    \centering
    \includegraphics[width=0.95\linewidth]{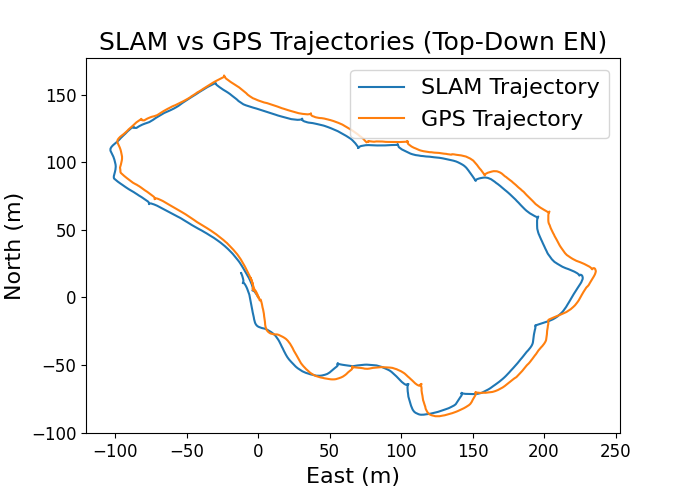}
    \caption{Without GNSS}
    \label{fig:flagstaff_no_gnss}
  \end{subfigure}\hfill
  \begin{subfigure}[t]{0.5\linewidth}
    \centering
    \includegraphics[width=0.95\linewidth]{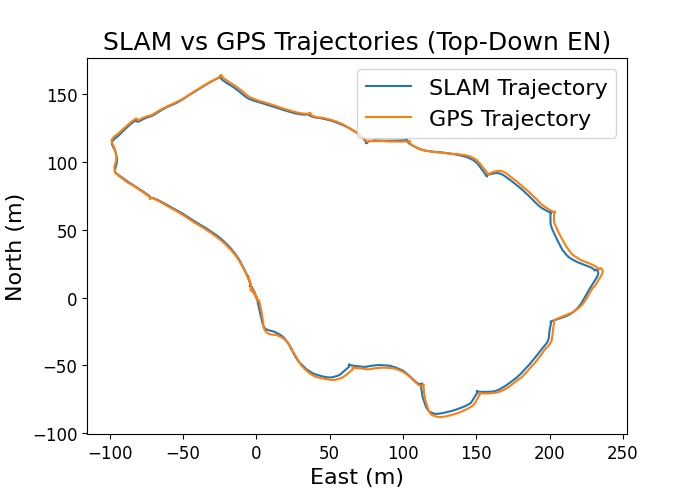}
    \caption{Covariance-aware GNSS}
    \label{fig:flagstaff_cov_aware}
  \end{subfigure}
  \caption{Reconstructed maps on the Flagstaff site for two GLIM variants.}
  \label{fig:flagstaff_maps}
\end{figure}

We conduct ablation studies to analyze SLAM robustness under degraded or absent GNSS. Specifically, we compare: (i) LiDAR-IMU SLAM without GNSS factors, (ii) GNSS integration with fixed covariance scaling (covariance-unaware), and (iii) covariance-aware GNSS integration. Table~\ref{tab:gnss_glim_variants1} reports the ATE between the estimated trajectory and RTK-GNSS ground truth for these variants. We observed that covariance-aware GNSS integration reduces trajectory error by 67.3\% relative to the no-GNSS baseline. Fig.~\ref{fig:flagstaff_maps} visualizes the estimated trajectories against RTK-GNSS ground truth, comparing GLIM with and without GNSS factors.
\begin{table}[h!]
  \centering
  \caption{SLAM trajectory error (m) for GLIM variants}
  \label{tab:glim_rmse}
  \vspace{-0.5em}
  \begin{tabular}{l c}
    \toprule
    \textbf{Method} & \textbf{Error (m)} \\
    \midrule
    Without GNSS                  & $8.67 \pm 4.77$ \\
    Constant-scaling GNSS         & $5.80 \pm 3.66$ \\
    Covariance-aware GNSS         & $2.83 \pm 2.51$ \\
    \bottomrule
  \end{tabular}
  \label{tab:gnss_glim_variants1}
  \vspace{-0.8em}
\end{table}

To test robustness to GNSS outliers, we synthetically corrupt the recorded GNSS stream with noise spikes ($\sim$10\,m), dropout windows ($\sim$15--30\,s), and constant-offset segments ($\sim$10\,m for $\sim$10\,s). These perturbations are designed to simulate controlled dense forest conditions, where multipath effects, canopy occlusions, and intermittent satellite visibility introduce large errors and signal loss. Table~\ref{tab:glim_robust_rmse} summarizes the resulting trajectory error. Incorporating the Huber loss improves performance over the baseline that uses a standard squared loss, indicating increased resilience to GNSS outliers.
\begin{table}[h!]
  \centering
  \caption{Effect of robust optimization on SLAM accuracy. Trajectory Error (m) for GLIM with and without a robust kernel}
  \label{tab:glim_robust_rmse}
  \vspace{-0.6em}
  \begin{tabular}{l c}
    \toprule
    \textbf{Method} & \textbf{Error (m)} \\
    \midrule
    Baseline (no robust kernel) & $2.28 \pm 2.24$ \\
    Robust kernel               & $1.95 \pm 1.88$ \\
    \bottomrule
  \end{tabular}
  \vspace{-0.9em}
\end{table}

\subsubsection{Tree of Heaven Detection}
\begin{figure}[h!]
    \centering
    \begin{subfigure}[t]{0.49\linewidth}
    \centering
    \includegraphics[width=\linewidth]{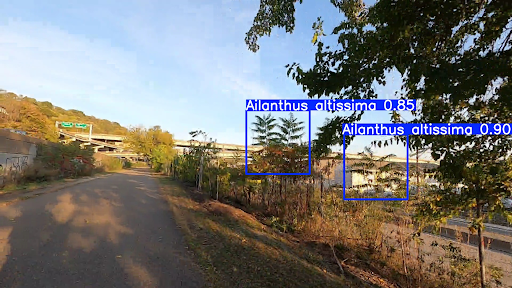}
  \end{subfigure}\hfill
  \begin{subfigure}[t]{0.49\linewidth}
    \centering
    \includegraphics[width=\linewidth]{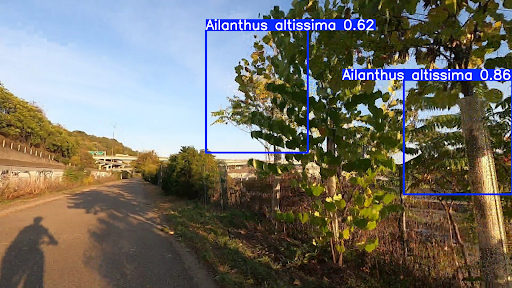}
    \end{subfigure}
    \caption{Tree-of-Heaven detections from our YOLOv8 model in a bike trail site, showing predicted bounding boxes and confidence scores for Ailanthus altissima.}
    \label{fig:yolo_examples}
\end{figure}

We evaluate the image-based Tree-of-Heaven (ToH) detection module using standard object detection metrics on the held-out test set. At IoU$=0.5$ with a confidence threshold of $0.35$, the detector achieves a precision of $0.615$, recall of $0.696$, F1-score of $0.653$, and mAP@0.5 of $0.686$. Qualitative examples are shown in Fig.~\ref{fig:yolo_examples}, illustrating detection performance on the Hazelwood bike trail site.

To better understand detection errors, we analyze false positives and false negatives by site. On the Hazelwood test split (31 images), the detector produced 60 predictions: 37 true positives (TP), 13 false positives (FP), and 10 false negatives (FN). Notably, all false positives correspond to other tree species (i.e., no non-tree background regions were spuriously detected as ToH). The missed ToH instances in Hazelwood are primarily due to occlusion by surrounding trees.

On the Gascola split, we observe a similar trend: false positives are predominantly confusions with visually similar trees. However, Gascola exhibits a higher rate of missed detections, largely attributable to reduced image quality, and truncated views where key cues are not visible.

\subsubsection{Aerial–Terrestrial Map Alignment Results}
We evaluate alignment accuracy on local map patches by computing the SE(2) error between a manually obtained ground-truth alignment and the transformation recovered by Tensor-MI. For each patch, we manually align the aerial and terrestrial maps, apply random SE(2) perturbations to the terrestrial point cloud with $t_x,t_y \in [-7,7]\mathrm{m}$ and $\psi \in [-30^\circ,30^\circ]$, and task each method with recovering the original alignment.
As summarized in Table~\ref{tab:map_alignment}, Tensor-MI achieves the lowest translation and yaw errors relative to ground truth, outperforming ICP, Fast Global Registration, and Clique~\cite{casseau2024markerless} even under significant canopy overlap and structural mismatch. 
Clique exhibits large yaw errors in our datasets due to spurious feature associations under heavy canopy overlap, though it performs better in sparser scenes where trunk/crown features are more distinctive.
\begin{figure}[h!]
  \centering
  \begin{subfigure}[t]{0.49\linewidth}
    \centering
    \includegraphics[width=0.7\linewidth]{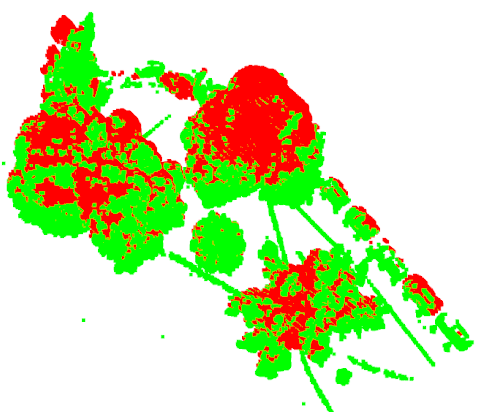}
    \caption{Flagstaff site}
    \label{fig:flagstaff_no_gnss}
  \end{subfigure}
  \begin{subfigure}[t]{0.49\linewidth}
    \centering
    \includegraphics[width=0.7\linewidth]{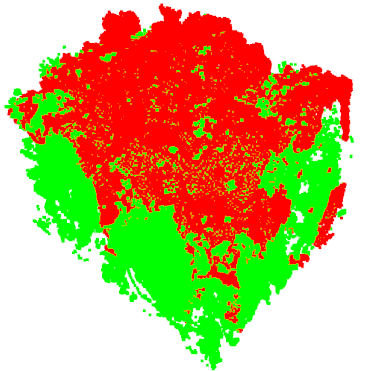}
    \caption{Chatham forest site}
    \label{fig:flagstaff_cov_aware}
  \end{subfigure}
  \caption{Aerial–terrestrial map alignment results on two patches. Green points show the terrestrial (under-canopy) map and red points show the aerial (above-canopy) map; the strong overlap indicates accurate alignment.}
  \label{fig:map_alignment}
\end{figure}

\begin{table}[h!]
  \centering
  \caption{Horizontal alignment error (translation in m, yaw in deg) on three patches.}
  \label{tab:map_alignment}
  \vspace{-0.6em}
  \setlength{\tabcolsep}{3.5pt}
  \renewcommand{\arraystretch}{1.05}
  \begin{tabular}{l cc cc cc}
    \toprule
    \textbf{Method} &
    \multicolumn{2}{c}{\textbf{Patch 1}} &
    \multicolumn{2}{c}{\textbf{Patch 2}} &
    \multicolumn{2}{c}{\textbf{Patch 3}} \\
    \cmidrule(lr){2-3}\cmidrule(lr){4-5}\cmidrule(lr){6-7}
    & \textbf{Trans} & \textbf{Yaw} & \textbf{Trans} & \textbf{Yaw} & \textbf{Trans} & \textbf{Yaw} \\
    \midrule
    ICP                     & 1.71 & 1.49  & 3.53  & 2.83  & 5.55  & 2.32 \\
    Fast Global Registration & 4.13 & 28.07 & 23.87 & 9.43  & 64.79 & 30.51 \\
    Clique                  & 6.00 & -53.54& 9.62  & -75.02& 9.37  & -81.78 \\
    Tensor-MI (our approach)              & 0.27 & 0.70 & 0.73  & 0.61 & 0.83 & 3.92 \\
    \bottomrule
  \end{tabular}
  \vspace{-0.9em}
\end{table}

\subsubsection{Geo-referenced outputs}
\begin{figure}[t!]
  \centering

  \begin{subfigure}[t]{0.49\linewidth}
    \centering
    \includegraphics[width=\linewidth, trim=0 0.5cm 0 0.5cm, clip]{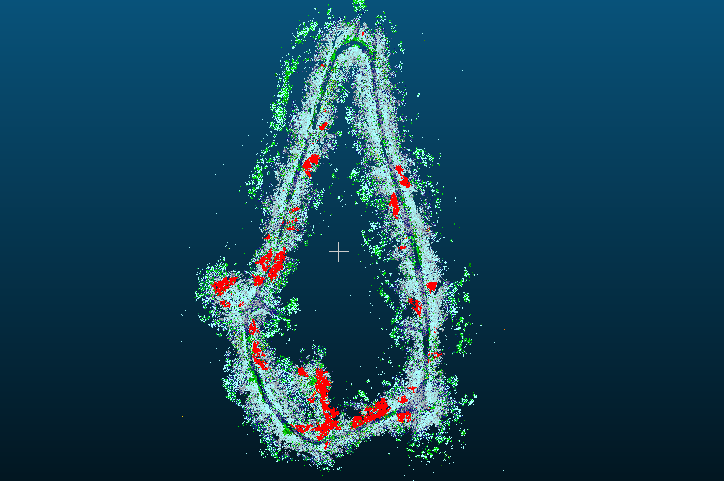}
    \label{fig:pc_gascola}
  \end{subfigure}\hfill
  \begin{subfigure}[t]{0.49\linewidth}
    \centering
    \includegraphics[width=\linewidth]{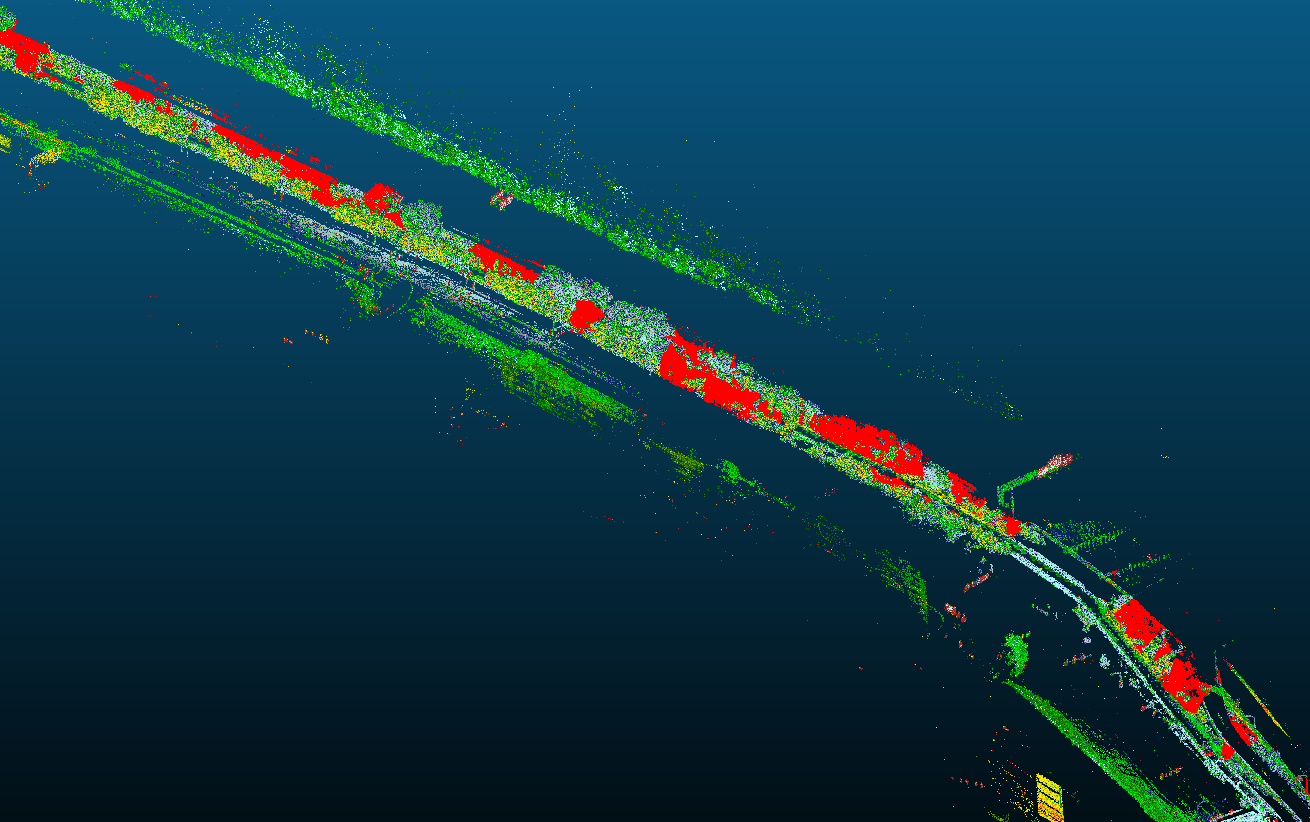}
    \label{fig:pc_biketrail}
  \end{subfigure}
  \begin{subfigure}[t]{0.49\linewidth}
    \centering
    \includegraphics[width=\linewidth, trim=0 2.5cm 0 2cm, clip]{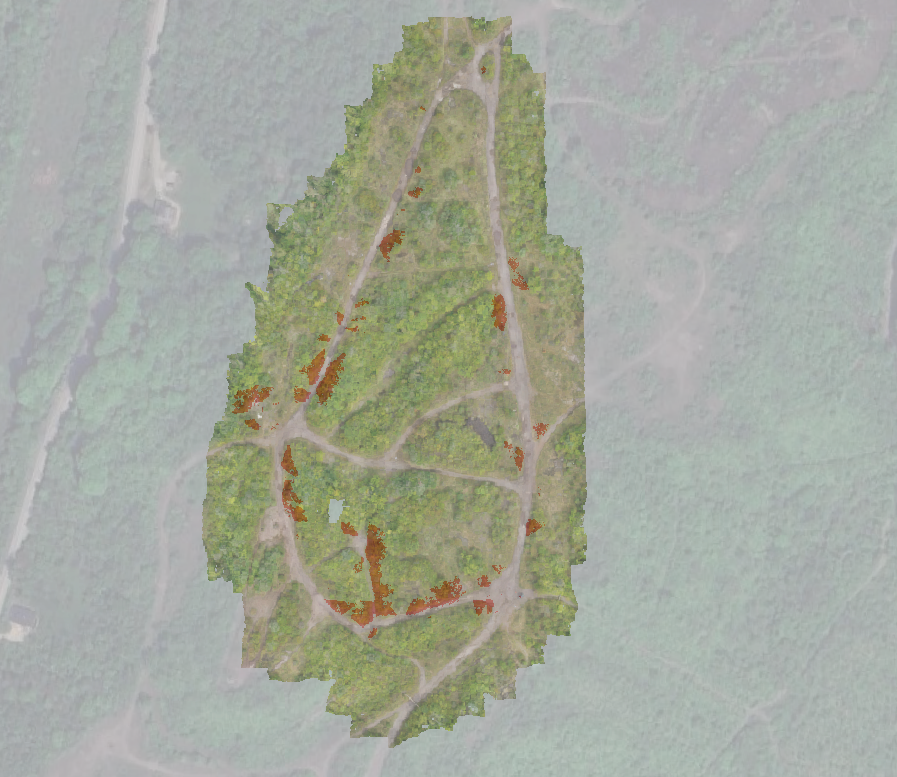}
  \end{subfigure}\hfill
  \begin{subfigure}[t]{0.49\linewidth}
    \centering
    \includegraphics[width=\linewidth, trim=0 0cm 0 1.2cm, clip]{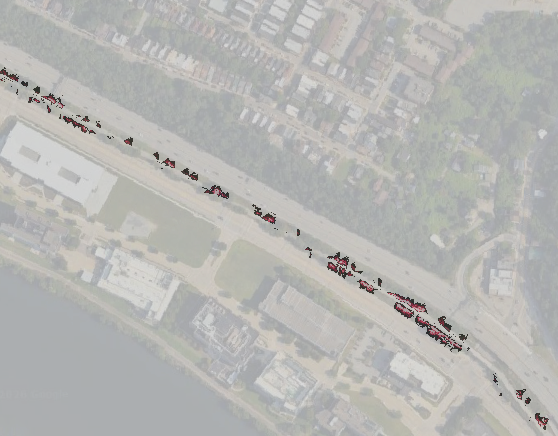}
    \label{fig:geotiff_bike}
  \end{subfigure}
  
  \caption{Geo-referenced Tree-of-Heaven outputs. Top: Point-cloud maps with detections highlighted in red. Bottom: GeoTIFF detection layers overlaid on Google Earth imagery.}
  \label{fig:toh_outputs}
  \vspace{-0.6em}
\end{figure}

We report geo-referenced outputs in two complementary forms (Fig.~\ref{fig:toh_outputs}). First, we generate georeferenced 3D point-cloud maps where Tree-of-Heaven detections are highlighted in red. Second, we export GIS-compatible GeoTIFF raster layers indicating detection regions. These layers can be visualized in standard GIS tools (e.g., QGIS) by overlaying them on satellite imagery. For the Gascola site, we additionally overlay the aerial reconstruction on the same basemap to illustrate consistent geospatial grounding across sensing modalities.

\section{Conclusion}
MapForest is a modular field robotics system for forest-scale mapping and invasive-species monitoring that converts multi-modal sensing (LiDAR, IMU, GNSS, and RGB) into georeferenced 3D reconstructions and GIS-ready geospatial layers highlighting Tree-of-Heaven. The system combines a compact, platform-agnostic sensing payload with a software pipeline that couples LiDAR--inertial SLAM (GLIM) and RGB-based detection, and grounds image-space detections within the reconstructed map to produce georeferenced outputs suitable for practitioner workflows. To improve long-range consistency in forested settings where under-canopy GNSS is noisy and intermittent, we integrated covariance-aware GNSS priors with robust loss functions into the mapping backend. 

There are several directions for future work. First, we plan to expand evaluation across additional seasons, canopy conditions, and geographic regions. Second, we plan to incorporate richer GIS maps, including per-tree attributes and uncertainty-aware map layers, to better support practitioner workflows. Finally, we will continue developing aerial--terrestrial alignment to merge above-canopy and under-canopy reconstructions into more complete forest representations. We will release the datasets and tooling to support reproducible research in forest mapping and invasive-species monitoring.

\section*{Acknowledgments}
This research was supported by the Richard King Mellon Foundation (Award No. AWD00002859). The authors also gratefully acknowledge the support of Chatham University’s Eden Hall Campus for assistance with data collection, as well as the members of the Kantor Lab at Carnegie Mellon University for their continued support and valuable input.
\bibliographystyle{IEEEtran}
\bibliography{references}


\end{document}